\documentclass[letterpaper, 10 pt, journal, twoside]{IEEEtran}

\ifCLASSINFOpdf
 
\else
  
\fi

\usepackage{graphicx} 
\usepackage{epsfig} 
\usepackage{mathptmx} 
\usepackage{times} 
\usepackage{amsmath}
\usepackage{amssymb}  

\usepackage{algorithm}
\usepackage{algorithmic}
\usepackage{multirow} 
\usepackage[caption=false,font=normalsize,labelfont=sf,textfont=sf]{subfig}
\usepackage{xcolor}
\usepackage{stfloats}

\newcommand{\cmmnt}[1]{}

\begin{document}

%
© 2025 IEEE. Personal use of this material is permitted.
Permission from IEEE must be obtained for all other uses,
including reprinting/republishing this material for advertising
or promotional purposes, collecting new collected works
for resale or redistribution to servers or lists, or reuse of
any copyrighted component of this work in other works.
This work has been submitted to the IEEE for possible
publication. Copyright may be transferred without notice,
after which this version may no longer be accessible.

\title{\color{black} Deep Reinforcement Learning-Based Motion Planning and PDE Control for Flexible Manipulators \color{black}}

\author{Amir Hossein~Barjini,
        Seyed Adel~Alizadeh Kolagar, Sadeq~Yaqubi,
        and~Jouni~Mattila
\thanks{The authors are with the Department of Engineering and Natural Sciences,
Tampere University, 7320 Tampere, Finland (e-mails: amirhossein.barjini@tuni.fi, seyedadel.alizadehkolagar@tuni.fi, sadeq.yaqubi@tuni.fi, jouni.mattila@tuni.fi). (Corresponding author: Amir Hossein Barjini.)}
\thanks{
This work was supported by the Research Council of Finland under the Project "Nonlinear PDE-model-based control of flexible manipulators" under Grant 355664.
}
}

\markboth{}%
{ }
%



\maketitle

\begin{abstract}
This \color{black} article \color{black} presents a motion planning and control framework for flexible robotic manipulators, integrating deep reinforcement learning (DRL) with a nonlinear partial differential equation (PDE) controller. Unlike conventional approaches that focus solely on control, we demonstrate that the desired trajectory significantly influences endpoint vibrations. To address this, a DRL motion planner, trained using the soft actor-critic (SAC) algorithm, generates optimized trajectories that inherently minimize vibrations. The PDE nonlinear controller then computes the required torques to track the planned trajectory while ensuring closed-loop stability using Lyapunov analysis. The proposed methodology is validated through both simulations and real-world experiments, demonstrating superior vibration suppression and tracking accuracy compared to traditional methods. The results underscore the potential of combining learning-based motion planning with model-based control for enhancing the precision and stability of flexible robotic manipulators.
\end{abstract}


\IEEEpeerreviewmaketitle

\section{Introduction}

Manipulators are widely used in robotic operations and industrial automation. Low energy consumption and lightweight design are the primary reasons for the growing popularity of flexible robotic manipulators. Despite these advantages, the inherent flexibility of these robots complicates modeling and control \cite{alandoli2020critical, chen2022variable, hu2019designing}. In general, manipulator flexibility refers to either flexible links \cite{sayahkarajy2016review} or flexible joints \cite{iskandar2022model}. 

\color{black} \subsection{Related Works} \color{black}

A review of the literature reveals various approaches for controlling flexible manipulators, \color{black}including hardware-based methods such as actuation via cables \cite{tang2021dynamic}, and software-based strategies like visual servoing \cite{li2021visual}\color{black}. For vibration suppression, some studies have modeled a fixed-free Euler-Bernoulli beam and aimed to control \color{black} the \color{black}endpoint vibrations using an input force \cite{jing2022backstepping, he2015vibration}. However, since it is important to simultaneously track the desired position and suppress vibrations in flexible manipulators, some studies have proposed boundary observer-based control with Lyapunov stability guarantees, applying a force for vibration suppression and a torque for tracking. Despite achieving good results in vibration suppression, these \color{black} methods \color{black} are limited to simulations, and actuation constraints remain a challenge for implementing input force in real-world experiments \cite{zhao2019boundary, jiang2015boundary}. An alternative, experimentally feasible approach is to use a single input torque applied to control the system's angular position \textcolor{black}{\cite{barjini2024deep}}. However, controlling endpoint vibrations in such an under-actuated system poses greater challenges \cite{viswanadhapalli2024deep}.

\color{black} Various \color{black} approaches have been adopted to tackle these challenges. Considering control strategies, some studies have utilized partial differential equations (PDEs) to design controllers \cite{yaqubi2023semi,sayyaadi2021boundary}, while others have transformed the PDEs into ordinary differential equations (ODEs), using methods such as assumed mode method (AMM), and designed controllers based on the simplified equations \cite{gao2018neural,he2020reinforcement}. Although designing a controller based on a reduced-order ODE model results in simplified controller design, it can negatively affect the precision of the designed controller at higher frequencies \cite{barjini2024design}.

From the motion planning perspective, extensive research has been conducted to optimize manipulator trajectories. However, most studies assume that manipulators have rigid-body links \cite{tamizi2023review, kroemer2021review}. Although the motion of flexible manipulators significantly influences system vibrations, there is still a lack of comprehensive research addressing this aspect.
The advancement of artificial intelligence has led to the emergence of learning-based methods for manipulator motion planning in unknown environments. Among these, reinforcement learning (RL) has become a prominent approach \cite{tamizi2023review}, enabling agents to interact with their environment, encounter diverse scenarios, and learn optimal actions through reward-based mechanisms \cite{sutton1998reinforcement}.
In recent years, with the progress of deep learning, traditional RL has evolved into deep reinforcement learning (DRL) \cite{del2022review}. By leveraging deep neural networks, DRL can handle high-dimensional, continuous state and action spaces, allowing robots to tackle more complex tasks such as navigation, adaptive motion planning, and task automation \cite{gu2017deep, prianto2021deep}.
\textcolor{black}{Among the various DRL algorithms, the Soft Actor-Critic (SAC) algorithm stands out for its superior performance in high-dimensional motion planning problems~\cite{haarnoja2018soft}. By incorporating an entropy term into its objective function, SAC enhances exploration and improves learning efficiency, enabling it to outperform existing methods in complex planning tasks \cite{prianto2021deep}. This could be especially useful for flexible-link manipulators, where nonlinearities and underactuation require broad exploration. SAC’s stochastic policy exploits passive dynamics, and its entropy maximization ensures robust, stable learning in complex systems, making it ideal for dynamic, uncertain environments.}

\textcolor{black}{
In recent years, DRL has been widely explored for motion planning in rigid-body manipulators, \textcolor{black}{with approaches ranging from end-to-end frameworks that directly output joint torques \cite{levine2016end}} to methods that combine DRL with low-level controllers \cite{shahna2024integrating, aljalbout2024role}.
However, to the best of our knowledge, the application of DRL to motion planning for flexible-link manipulators remains largely unexplored.}

\color{black} \subsection{Contributions} \color{black}

The main contributions of this paper can be summarized as follows:
1) Development of a nonlinear PDE controller for flexible manipulators that ensures closed-loop stability via Lyapunov theory, enabling precise trajectory tracking using only a single input torque at the base.
2) Introduction of a DRL motion planner for flexible manipulators, leveraging the SAC algorithm to generate optimal trajectories that actively suppress endpoint vibrations.
\textcolor{black}{3) Validation of the proposed synergistic model-based PDE and model-free DRL approach, effectively handling underactuation without endpoint actuation unlike traditional boundary control methods.}
4) Validation of the proposed motion planning and control framework through simulations and real-world experiments on a hydraulically actuated flexible robotic manipulator, demonstrating superior performance compared to conventional control and motion planning methods.

\color{black} This paper is organized as follows: \color{black}
Section II presents the mathematical modeling of the flexible manipulator, including the governing PDEs.  
In Section III, the proposed nonlinear PDE controller is introduced, followed by the development of the DRL motion planner.  
Section IV evaluates the proposed method through numerical simulations. 
Section V provides experimental validations on a real flexible robotic manipulator to verify the proposed method's effectiveness. Both the PDE controller and DRL motion planner are tested under real-world conditions.
Finally, Section VI summarizes the key findings and explores future research directions.

\section{Problem Formulation}
\begin{figure}[ht]
\centering
\includegraphics[width=0.3\linewidth]{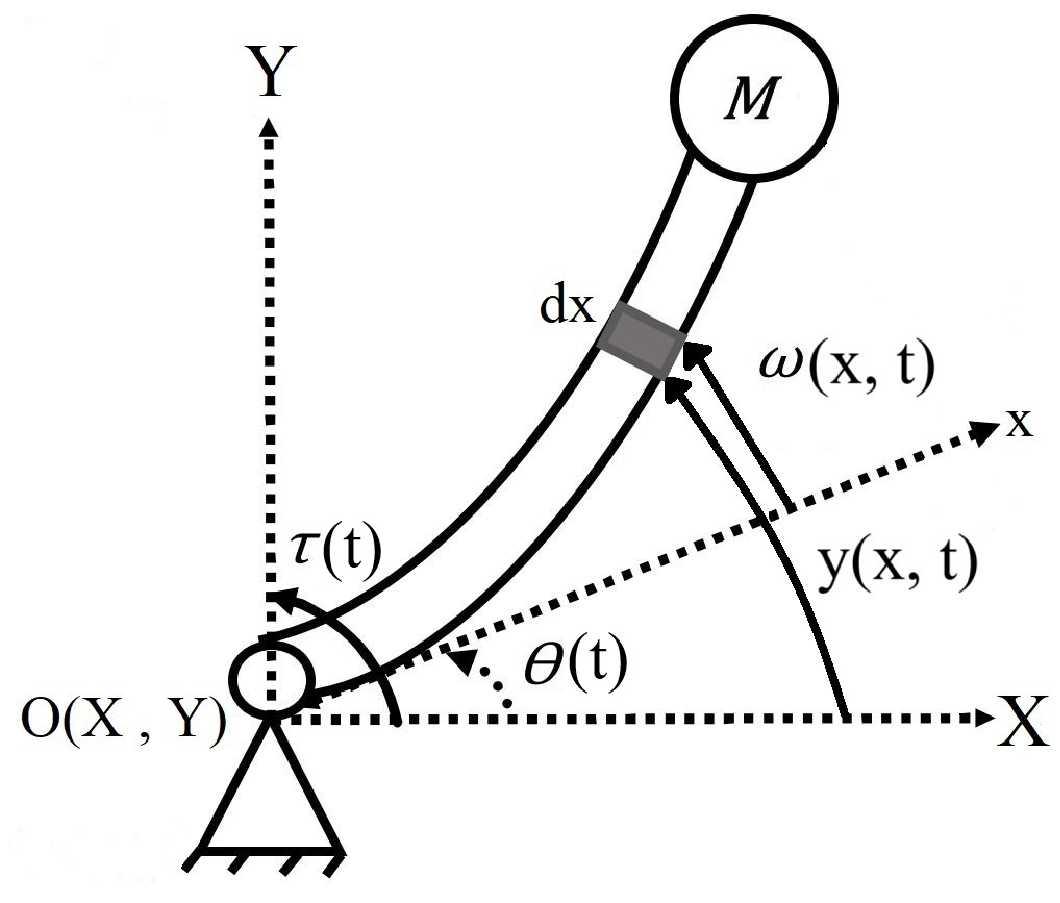}
\caption{Schematic View of the Flexible Manipulator.}
\label{Schematic Platform}
\end{figure}

In this chapter, we derive the PDE model for the flexible robotic manipulator. This model forms the foundation for both the model-based controller and the design of the motion planner for the system. Consequently, the accuracy of the model plays a crucial role in the reliability of the experimental results. The manipulator, a single-link system, operates in the vertical plane, where the system and payload's weight make gravity an influential factor. A schematic representation of the flexible manipulator is shown in Fig. \ref{Schematic Platform}.

The system is modeled as an infinite-dimensional system using Euler–Bernoulli beam theory, \color{black} where \( XOY \) and \( xOy \) denote the inertial and rotating coordinate systems, respectively. The variables \( M \), \( \theta \), \( \omega \), and \( \tau \) represent the mass of the payload, the angular position of the flexible link, the elastic deviation along the link, and the input torque, respectively. \color{black} Assuming small deformations, the arc position of each segment of the flexible link is expressed as $y(x, t)=x\theta(t)+\omega(x, t)$. \color{black} Throughout this paper, a dot denotes a time derivative (\( \dot{[-]}=\frac{\partial[-]}{\partial t} \)), while a prime represents a spatial derivative ( \( [-]'=\frac{\partial[-]}{\partial x} \) ). \color{black}

To derive the equations of motion, the kinetic energy ($T$), potential energy ($U$), and virtual work done by external torque ($W$), can be expressed as:

\begin{equation}
T= \frac{1}{2} I_m \dot \theta^2(t) + \frac{1}{2} \rho A \int_0^L \dot y^2(x, t)  dx + \frac{1}{2} M \dot y^2(L, t), \label{T}
\end{equation}
\begin{equation}
\begin{split}
U=& \frac{1}{2} mgL \sin{\theta(t)} + Mg(L\sin{\theta(t)} + \omega(L, t)\cos{\theta(t)})\\
+& \frac{1}{2} EI \int_0^L [\omega''(x, t)]^2 dx\\ 
=& \frac{1}{2} mgL \sin{\theta(t)} + MgL\sin{\theta(t)} \\ +& Mg(y(L, t)-L\theta(t))\cos{\theta(t)}
+ \frac{1}{2} EI \int_0^L [y''(x, t)]^2 dx,
\end{split}
\label{V}
\end{equation}
\begin{equation}
W=\tau(t) \theta(t), \label{W}
\end{equation}
\color{black} where \( I_m \), \( \rho \), \( A \), \( m \), \( L \), and \( EI \) denote the moment of inertia, density, cross-sectional area, mass, length, and bending stiffness of the flexible link, respectively. \color{black}

Now, applying the extended Hamilton's principle, as $ \int_{t_1}^{t_2} (\delta T - \delta U +\delta W) dt = 0 $, the governing equations of motion and the corresponding boundary conditions are derived as follows:

\begin{equation}
I_m \Ddot{\theta}(t) - EI \omega''(0, t) + \frac{1}{2} mgL \cos{\theta(t)} - Mg \omega(L, t)\sin{\theta(t)} = \tau(t),
\label{Equation of motion 1}
\end{equation}
\begin{equation}
\rho Ax \Ddot{\theta}(t) + \rho A \Ddot{\omega}(x, t) + EI \omega''''(x, t) = 0,
\label{Equation of motion 2}
\end{equation}
\begin{equation}
ML \Ddot{\theta}(t) + M \Ddot{\omega}(L, t) - EI \omega'''(L, t) + Mg \cos{\theta(t)}= 0,
\label{Equation of motion 3}
\end{equation}
\begin{equation}
\omega(0, t) = \omega'(0, t) = \omega''(L, t) = 0 .
\label{Equation of motion 4}
\end{equation}

Equations (\ref{Equation of motion 1}) and (\ref{Equation of motion 2}) represent the governing equations of motion, while equations (\ref{Equation of motion 3}) and (\ref{Equation of motion 4}) define the boundary conditions.

It can be observed that due to the gravity term $Mg \cos{\theta}$, the boundary condition (\ref{Equation of motion 3}) is non-homogeneous. Therefore, to obtain an accurate semi-analytical solution, a model transformation is required. To accomplish this, we first substitute (\ref{Equation of motion 2}) into (\ref{Equation of motion 3}) at $x=L$, resulting in the following modified boundary condition:

\begin{equation}
\omega''''(L, t) + p\omega'''(L, t) = f,
\label{New Boundary Condition}
\end{equation}
where $p=\frac{\rho A}{M}$ and \color{black} $f=\frac{\rho A g}{EI} \cos{\theta}$ \color{black}. To homogenize the boundary conditions (\ref{Equation of motion 4}) and (\ref{New Boundary Condition}), we apply the following transformation, which maps $\omega(x, t)$ to $z(x, t)$, as follows \cite{yaqubi2023semi}:

\begin{equation}
z(x, t)=\omega(x, t)+\nu(x, t).
\label{Z}
\end{equation}

$\nu(x, t)$ is chosen such that $z(x, t)$ satisfies the following homogeneous boundary conditions:

\begin{equation}
z(0, t) = z'(0, t) = z''(L, t) = 0,
\label{New Boundary Z 1}
\end{equation}
\begin{equation}
z''''(L, t) + pz'''(L, t) = 0.
\label{New Boundary Z 2}
\end{equation}

As a result, for $\nu(x, t)$ we should have:

\begin{equation}
\nu(0, t) = \nu'(0, t) =\nu''(L, t) = 0,
\label{Nu Conditions 1}
\end{equation}
\begin{equation}
\nu''''(L, t) + p\nu'''(L, t) = -f.
\label{Nu Conditions 2}
\end{equation}

Solving equations (\ref{Nu Conditions 1}) and (\ref{Nu Conditions 2}) yields the following expression for $\nu(x)$:
\begin{equation}
\nu(x, t)=\color{black}\left( \color{black}-\frac{\Gamma_1}{p^4}e^{-px}+\frac{1}{6p}x^3+\Gamma_2 x^2-\frac{\Gamma_1}{p^3}x+ \frac{\Gamma_1}{p^4} \color{black} \right) \color{black} f
\label{Nu}
\end{equation}
where $\Gamma_1$ and $\Gamma_2$ are defined as:
\begin{equation}
\Gamma_1= \frac{2p^3L^3}{(3p^2L^2-6)e^{-pL}+6-6pL},
\label{Gamma 1}
\end{equation}
\begin{equation}
\Gamma_2= \frac{\Gamma_1}{2p^2}e^{-pL}-\frac{1}{2p}.
\label{Gamma 2}
\end{equation}

Now, with $\nu(x, t)$ obtained from (\ref{Nu}), the homogenized parameter $z(x, t)$ can be solved using the Assumed Mode Method (AMM).

\section{Proposed Methodology}

In the previous chapter, it was shown that the flexible manipulator is controlled using a single input torque, which is responsible for adjusting the angular position ($\theta$) of the system. Therefore, due to the flexibility of the robot, the system has infinite degrees of freedom and is considered under-actuated. This makes it challenging to control all degrees of freedom using only one control input. For example, while the controller can effectively control the angular position ($\theta$), it may still result in undesired vibrations ($\omega$) during sharp motions. In fact, both the controller and the motion planner play critical roles in this system. To effectively suppress vibrations and reach the target as efficiently as possible, it is essential to derive an optimal trajectory between each angle.

\begin{figure}[ht]
\centering
\includegraphics[width=0.5\linewidth]{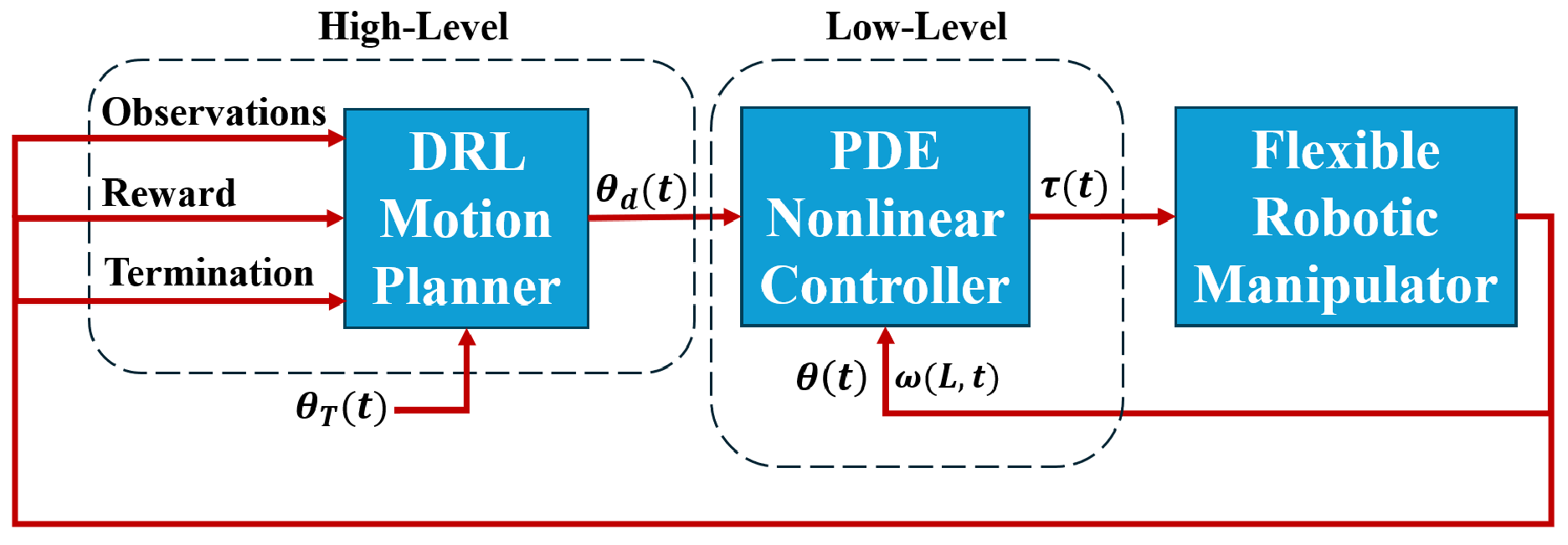}
\caption{Schematic View of the Proposed High-Level Motion Planner and the Low-Level Controller.}
\label{Schematic Methodology}
\end{figure}

In this study, we propose a novel approach that combines a motion planner using DRL with a nonlinear PDE controller. In this high-level-low-level framework, the high-level motion planner first generates the desired angular position ($\theta_d$) to reach the target angle ($\theta_T$), in a manner that minimizes vibrations. The low-level controller, consisting of the PDE nonlinear controller, then ensures the tracking of this desired angular position by producing the required input torque, as shown in Fig. \ref{Schematic Methodology}.

\subsection{Nonlinear PDE Controller}
As observed in the previous section, the input torque $\tau$ is applied to the system to control the angular position of the manipulator. In this section, a nonlinear PDE controller is proposed to control the flexible manipulator, which guarantees the stability of the system. \textcolor{black}{The primary motivation for using a PDE controller lies in its model-based nature, which is particularly beneficial for systems with distributed parameters, such as flexible links. Since the dynamics of flexible structures are inherently described by partial differential equations, incorporating PDEs into the controller design allows for a more accurate representation of the system behavior, as follows:} 

\begin{equation}
\tau(t) = K_p e(t) + K_d \dot{e}(t) - H(t), \label{Tau}
\end{equation}
\begin{equation}
\begin{split}
H(t) =& EI \omega''(0, t) - \frac{1}{2} mgL \cos{\theta(t)} \\+& Mg \omega(L, t)\sin{\theta(t)} - I_m \ddot{\theta}_d(t), \label{H}
\end{split}
\end{equation}
where $e(t) = \theta_d(t) - \theta(t)$ represents the tracking error, and $K_p$, $K_d$, and $\alpha$ are positive parameters. Consider the following Candidate Lyapunov Function (CLF), defined as:
\begin{equation}
V = \frac{1}{2} K_p e^2 + \frac{1}{2} I_m \dot{e}^2 + \alpha I_m e \dot{e}. \label{CLF}
\end{equation}

\textcolor{black}{In this work, the Lyapunov function includes a coupling term ($\alpha I_m e \dot{e}$) to enable designing a controller with provable exponential stability. This term allows the derivative to satisfy $\dot{V} < -\lambda V$, ensuring exponential error decay.} 

\textbf{Lemma 1:}
By selecting appropriate parameters for $\alpha$, $I_m$, and $K_p$ with the conditions $\alpha I_m < K_p$ and $\alpha < 1$, the proposed CLF in equation (\ref{CLF}) \color{black} is \color{black} a positive function. 

\textbf{Proof}:

We begin by considering the following inequality:

\begin{equation}
(e \pm \dot{e})^2 = e^2 + \dot{e}^2 \pm 2 e \dot{e} \geq 0, \label{Proof 1_0}
\end{equation}
thus, if we multiply both sides of this inequality by the positive value $\alpha I_m$, we obtain:

\begin{equation}
|\alpha I_m e \dot{e}| \leq \frac{1}{2} \alpha I_m (e^2 + \dot{e}^2), \label{Proof 1_1}
\end{equation}
so, if we select $\alpha$ and $K_p$ such that $\alpha I_m < K_p$ and $\alpha < 1$, we can conclude:

\begin{equation}
|\alpha I_m e \dot{e}| \leq \frac{1}{2} \alpha I_m (e^2 + \dot{e}^2) \leq \frac{1}{2} K_p e^2 + \frac{1}{2} I_m \dot{e}^2, \label{Proof 1_2}
\end{equation}
therefore, considering the proposed CLF (\ref{CLF}) and the inequality in (\ref{Proof 1_2}), we conclude that $V \geq 0$.

\textbf{Lemma 2:}
By assuming a condition for $\alpha$, $I_m$ and $K_d$  as $\alpha I_m < K_d$, the time derivative of the proposed CLF \color{black} is \color{black} upper bounded.

\textbf{Proof:}
\begin{equation}
\begin{split}
\dot{V} &= K_p e \dot{e} + I_m \dot{e} \ddot{e} + \alpha I_m (\dot{e}^2 + e \ddot{e})\\
&= (\dot{e}+ \alpha e) I_m \ddot{e} + K_p e \dot{e} + \alpha I_m \dot{e}^2 \\
&= (\dot{e} + \alpha e) [EI \omega''(0, t) - \frac{1}{2} mgL \cos{\theta} + Mg \omega(L)\sin{\theta} \\
&+ \tau(t) - I_m \ddot{\theta}_d] + K_p e \dot{e} + \alpha I_m \dot{e}^2,
\end{split}
\label{Proof 2_1}
\end{equation}
now, by substituting the control law (\ref{Tau}) into equation (\ref{Proof 2_1}), we have:

\begin{equation}
\begin{split}
\dot{V} &= -\alpha K_p e^2 - K_d \dot{e}^2 - K_d \alpha e \dot{e} + \alpha I_m \dot{e}^2 \\
&=-\alpha K_p e^2 - (K_d - \alpha I_m) \dot{e}^2 - K_d \alpha e \dot{e}. 
\end{split}
\label{Proof 2_2}
\end{equation}

As $\alpha I_m < K_d$, we can define the following three positive variables $\lambda_1$, $\lambda_2$, and $\lambda_3$:

\begin{equation}
\alpha K_p = \lambda_1 \color{black}\left(\color{black} \frac{1}{2} K_p \color{black} \right) \color{black},
\label{lambda1}  
\end{equation}
\begin{equation}
K_d - \alpha I_m = \lambda_2 \color{black}\left(\color{black} \frac{1}{2} I_m \color{black} \right) \color{black},
\label{lambda2}  
\end{equation}
\begin{equation}
\alpha K_d = \lambda_3 ( \alpha I_m ).
\label{lambda3}  
\end{equation}

Consequently, we can write equation (\ref{Proof 2_2}) as follows:

\begin{equation}
\dot{V} = -\lambda_1 \left(\frac{1}{2} K_p e^2 \right) - \lambda_2 \left(\frac{1}{2} I_m \dot{e}^2 \right) - \lambda_3 \left(\alpha I_m e \dot{e} \right). 
\label{Proof 2_3}
\end{equation}

Now, let's define $\lambda$ as it satisfies the following condition:

\begin{equation}
\lambda < min\{\lambda_1, \lambda_2, \lambda_3\},
\label{lambda}  
\end{equation}

Thus, by selecting the parameter $\lambda$ that satisfies the condition (\ref{lambda}), we conclude that:

\begin{equation}
\dot{V} < -\lambda \left(\frac{1}{2} K_p e^2 + \frac{1}{2} I_m \dot{e}^2 +\alpha I_m e \dot{e} \right) = -\lambda V. 
\label{Proof 2_4}
\end{equation}

\textbf{Theorem 1:}
The control law (\ref{Tau}) ensures that the system tracks the desired angle ($\theta_d$) and guarantees the exponential stability of the closed-loop system.

\textbf{Proof:}

As shown in Lemma 1 and Lemma 2, $V \geq 0$ and $\dot{V} < -\lambda V$, therefore, from (\ref{Proof 2_4}), we have:

\begin{equation}
\frac{dV}{V} < -\lambda dt  , 
\label{Stability Proof_2}
\end{equation}
By integrating this inequality over $[0, t]$, we can obtain:
\begin{equation}
V(t) < e^{-\lambda t} V(0), 
\label{Stability Proof_1}
\end{equation}
Therefore, according to the Lyapunov stability theorem, the closed-loop system will be exponentially stable, and as $t \to \infty$, the system will track the desired angular position, leading to $e \to 0$.

\subsection{Deep Reinforcement Learning for Motion Planning}

As explained previously, both the controller and the motion planner play a crucial role in vibration suppression of flexible manipulators. We propose a DRL agent that generates motion plans for a flexible robotic manipulator to minimize vibrations while achieving the target angle. The SAC algorithm, a model-free DRL approach, has demonstrated high effectiveness in motion planning tasks \cite{prianto2021deep}.
Unlike standard RL, which aims to maximize the expected cumulative reward $\sum_t \mathbb{E}_{(s_t,a_t) \sim \rho_\pi} [r(s_t,a_t)]$,
the SAC algorithm optimizes a more general objective that incorporates entropy regularization. This objective encourages exploration by promoting stochastic policies, formulated as:
\begin{equation}
 J(\pi) = \sum_t \mathbb{E}_{(s_t,a_t) \sim \rho_\pi} [r(s_t,a_t) + \beta H(\pi(\cdot | s_t))],
\label{Entropy}
\end{equation}
where \( H(\pi(\cdot | s_t)) \) represents the entropy of the policy, and \( \beta \) is a temperature parameter that controls the trade-off between reward maximization and policy stochasticity. In this algorithm, we use a soft state value function $V_\psi(s_t)$, two soft Q-functions $Q_{\theta_i}(s_t,a_t)$, and a policy $\pi_\phi(a_t|s_t)$, all implemented as neural networks with parameters $\psi$, $\theta_i$, and $\phi$, respectively\cite{haarnoja2018soft}.

\textcolor{black}{Building on established research, the state and action spaces, as well as the reward function, were carefully adapted to capture the unique characteristics of our problem, enabling stable learning and consistent convergence.}

In this context, the state, \( s_t \), is defined as follows:
\begin{equation}
\begin{aligned}
\label{states vector}
s_t=\left\langle e_T(t), \theta(t), \dot{\theta}(t), \tau(t), \omega(L, t) , \dot{\omega}(L, t)\right\rangle,
\end{aligned}
\end{equation}
where $e_T(t) = \theta_T(t) - \theta(t)$ represents the difference between the target joint angle $\theta_T$ and the current joint angle $\theta$, and $\tau$ is the input torque. \( \dot{\theta} \) and \( \dot{\omega} \) denote the time derivatives of \(\theta \) and \( \omega \), respectively, in time step \( t\).

\textcolor{black}{Joint velocities were used as the action space, $a_t$, as they are well-suited for motion planning tasks and have demonstrated superior reliability in sim-to-real transfer scenarios \cite{aljalbout2024role}. The continuous action vector is defined as:
\begin{equation}
\begin{aligned}
\label{action vector 1}
a_t \sim \pi_\phi(a_t|s_t),
\end{aligned}
\end{equation}
where $a_t \in [-1, 1]$ represents normalized joint velocity commands. These actions are then scaled by the maximum allowable joint velocity, $\dot{\theta}_{\max}$, to produce the final desired joint velocities:
\begin{equation}
\begin{aligned}
\label{action vector 2}
\dot{\theta}_d = \dot{\theta}_{\max} \cdot a_t,
\end{aligned}
\end{equation}
resulting in a final velocity range of $[-\dot{\theta}_{\max}, \dot{\theta}_{\max}]$. The value of $\dot{\theta}_{\max}$ is manually specified according to the physical characteristics and limitations of the manipulator.
} The desired angle \( \theta_d(t) \) is then obtained by integrating the desired angular velocity:
\begin{equation}
\begin{aligned}
\label{action vector 4}
\theta_d = \int_0^t \dot{\theta}_d dt.
\end{aligned}
\end{equation}

\textcolor{black}{The reward function is defined in Eq.~\eqref{Reward function} and consists of five terms, each designed to address specific challenges posed by the flexible-link manipulator.}
\begin{equation}
    \label{Reward function}
    R_t = W_e |e_T(t)| + W_{\dot{\theta}} |\dot{\theta}(t)| + W_{\dot{\omega}} |\dot{\omega}(L, t)| + R_{reach} + R_{failure}.
\end{equation}
\textcolor{black}{While error deviation alone is typically sufficient for standard reaching tasks, the additional terms were introduced to handle the unique dynamics of flexible manipulators. The first term, $W_e |e_T(t)|$, penalizes the deviation between the target and the current joint angle, which is common in reaching tasks. The second term, $W_{\dot{\theta}} |\dot{\theta}(t)|$, penalizes excessive joint velocity, a critical factor for flexible manipulators as high velocities can lead to instability. The third term, $W_{\dot{\omega}} |\dot{\omega}(L, t)|$, addresses tip vibration by penalizing excessive tip velocity, helping to reduce oscillations inherent in flexible manipulators. Additionally, a positive reward, $R_{reach}$, is given when the agent successfully reaches the target under desired conditions, while $R_{failure}$ imposes a penalty if the manipulator exceeds its operational limits. These terms were manually tuned to ensure convergence and a balanced exploration-exploitation trade-off, with their combination tailored to the specific demands of our flexible manipulator task.}


\color{black}As shown in Algorithm 1, the proposed method combines high-level model-free motion planning using SAC with low-level model-based controller to achieve accurate motion tracking for flexible manipulators while suppressing the vibration.

\begin{algorithm}[H]
\color{black} \caption{Motion Planning and Control for Flexible Manipulators}
\begin{algorithmic}[1]
\REQUIRE Target trajectory $\theta_T(t)$, initial states
\ENSURE Input torque $\tau(t)$ applied to the flexible manipulator

\STATE \textbf{Initialize}: SAC policy $\pi(s_t)$, PDE controller, initial conditions
\FOR{each time step $t$}
    \STATE \textbf{Motion Planning (High-Level)}:
    \STATE DRL State: $s_t \gets \langle e_T(t), \theta(t), \dot{\theta}(t), \tau(t), \omega(L, t), \dot{\omega}(L, t) \rangle$
    \STATE Generate desired velocity: $\dot{\theta}_d(t) \gets \pi(s_t)$
    \STATE Calculate the desired position: $\theta_d(t) \gets \int_0^t \dot{\theta}_d dt$
    
    \STATE \textbf{Control (Low-Level)}:
    \STATE Compute input torque (from PDE controller in equation (17)): $\tau(t) \gets f_{\text{PDE}}(\theta_d(t), \theta(t), \omega(L, t))$
    \STATE Apply torque $\tau(t)$ to the manipulator
\ENDFOR
\end{algorithmic}
\end{algorithm}
\color{black}

\section{Numerical Simulation}

Now, to evaluate the performance of the proposed high-level motion planner and low-level controller, a comparative analysis is conducted between traditional methods and the proposed hybrid method. The flexible manipulator is simulated in MATLAB Simulink, considering three modes. The simulation results for the conventional PID controller, the proposed PDE controller, and the combined PDE controller with the DRL motion planner are presented. The parameters used in these simulations are based on the real robot and are detailed in Table \ref{Parameters}.

\begin{table}[ht]
\caption{Parameters of the Flexible Manipulator and their Values}
\centering
\begin{tabular}{|c|c|c|}
\hline
\textbf{Symbol} & \textbf{Definition} & \textbf{Value (Unit)}\\ [0.5ex]
\hline
$L$ & Length of the Flexible Link & 4.5 ($m$)  \\ \hline
$\rho$ & Density of the Flexible Link & 7850 ($Kg/m^3$)  \\ \hline
$A$ & Cross Area of the Flexible Link &$6.84 \times 10^{-4}$ ($m^2$) \\ \hline
$E$ & Young's Modulus of the Flexible Link & 200 ($GPa$)  \\ \hline
$I$ & Moment of Inertia of the Flexible Link & $3.71 \times 10^{-7}$ ($m^4$)  \\ \hline
$M$ & Payload Mass & 20 ($Kg$) \\ \hline
\end{tabular}
\label{Parameters}
\end{table}

\subsection{Simulation Results Using the Conventional PID Controller}

To track the desired angular position ($\theta_d$), based on the cubic polynomial trajectory (CPT) method, we consider the conventional PID controller defined as:

\begin{equation}
\tau(t) = K_p e(t) + K_i \int^t_0{e(t)} dt + K_d \dot{e}(t), \label{Tau_PID}
\end{equation}
where, $e(t)$ is the tracking error, defined as $e(t) = \theta_d(t) - \theta(t)$, and $K_p$, $K_i$, and $K_d$ are the proportional, integral, and derivative gains, respectively. By implementing this PID controller with the selection of $K_p = 150000$, $K_i = 100000$, and $K_d = 20000$, the simulation results are presented in blue in Fig. \ref{Simulation PID PDE}.

As observed in Fig. \ref{Simulation PID PDE Theta}, although the result of the PID controller for tracking is acceptable, the vibrations at the payload, shown in Fig. \ref{Simulation PID PDE Omega}, are significant, especially when the flexible manipulator starts to move between angles. Additionally, large jumps in the required torque are observed in in Fig. \ref{Simulation PID PDE Torque}, which could potentially harm the actuator in a real robot.

\subsection{Simulation Results Using the Proposed PDE Controller}

\begin{figure}[H]
\centering
\subfloat[]{\includegraphics[width=1.3in]{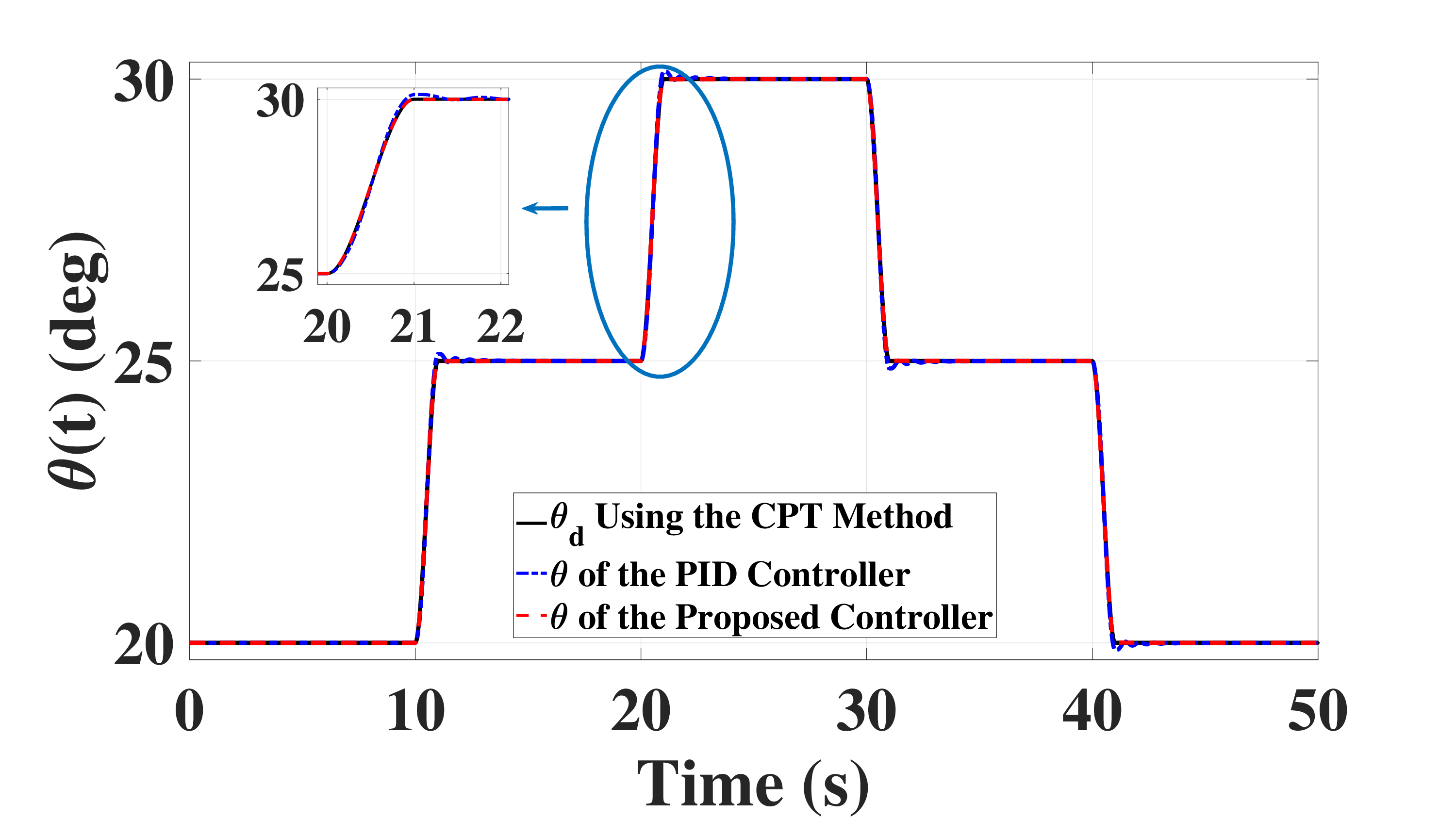}%
\label{Simulation PID PDE Theta}}

\subfloat[]{\includegraphics[width=1.3in]{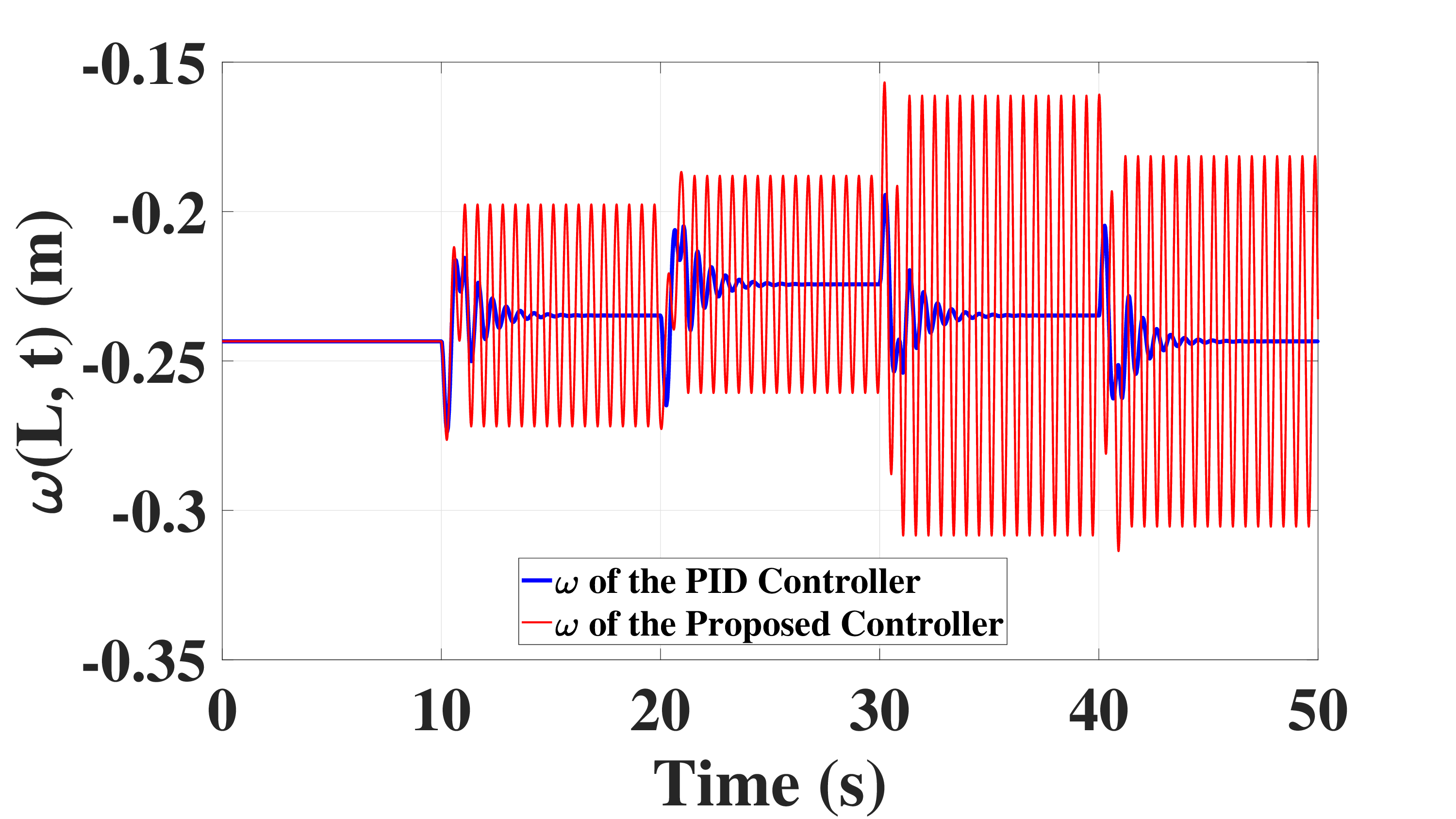}%
\label{Simulation PID PDE Omega}}
\subfloat[]{\includegraphics[width=1.3in]{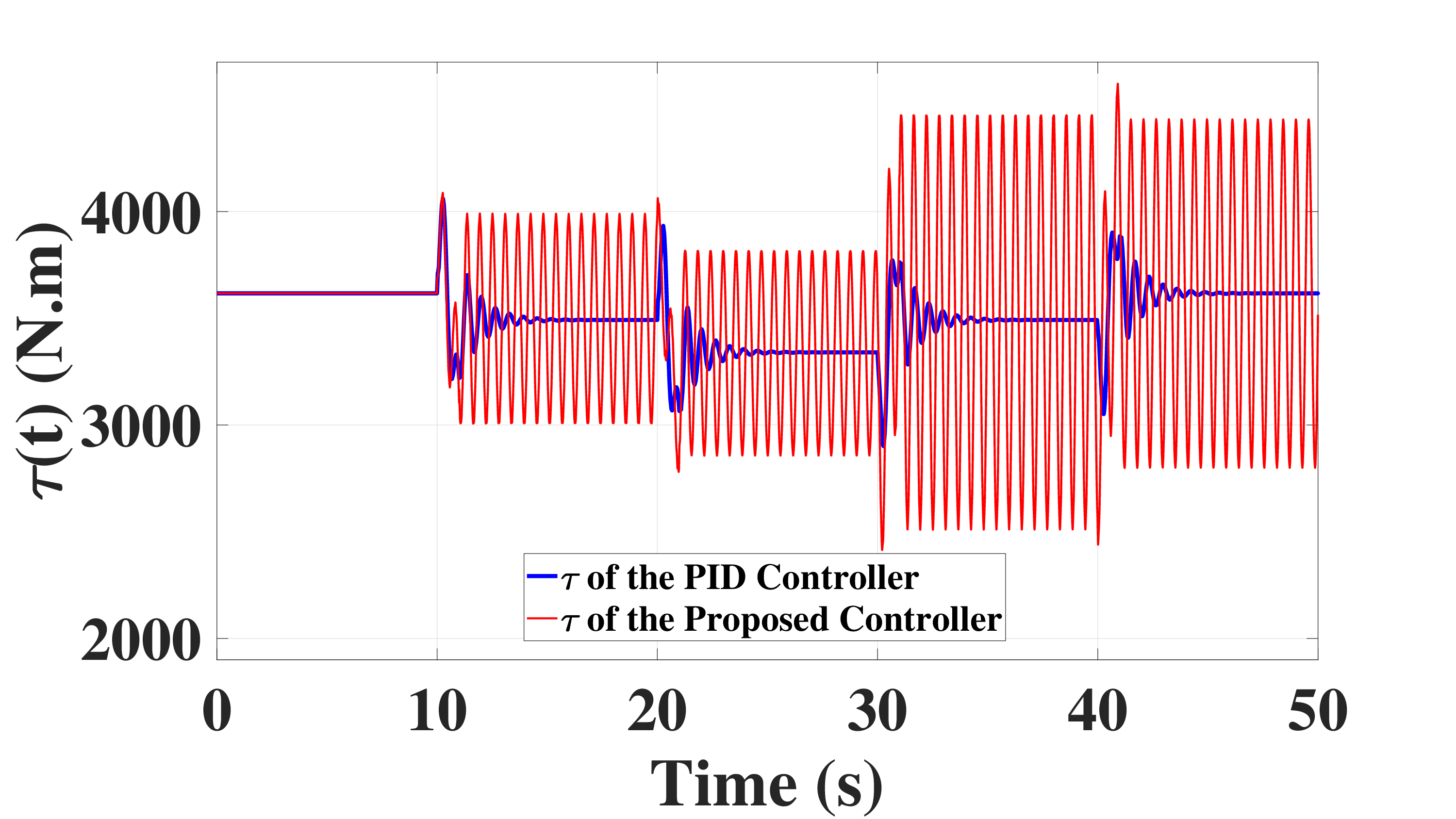}%
\label{Simulation PID PDE Torque}}
\caption{\textcolor{black}{Comparison of the PID Controller and the Proposed PDE Controller Using the CPT Method. (a) Angular Position. (b) Elastic Deviation of the Payload. (c) Control Input Torque.}}
\label{Simulation PID PDE}
\end{figure}

Now, by applying the proposed nonlinear PDE controller, as presented in equation (\ref{Tau}), with $K_p = 12000$ and $K_d = 15000$, the system is simulated as in the previous section. These simulation results are also depicted in red in Fig. \ref{Simulation PID PDE}, where the system is expected to track the same desired angular position ($\theta_d$) as the previous section, generated using the CPT method.
As observed in Fig. \ref{Simulation PID PDE Theta}, the tracking performance was significantly improved compared to the PID controller. Furthermore, it is important to note that while the conventional PID controller cannot guarantee the stability of the closed-loop system, the proposed controller ensures stability.

However, as shown in Fig. \ref{Simulation PID PDE Omega}, the proposed controller did not effectively suppress the vibrations. This can be attributed to two main reasons: First, the PDE controller focuses solely on accurately controlling the angular position ($\theta$) without addressing the suppression of payload vibrations. Second, the system is under-actuated, with only one control input; since there is no control input applied to the end effector, it cannot control the payload vibrations ($\omega(L, t)$). Thus, in this under-actuated system, utilizing a motion planner for vibration suppression becomes essential.

Furthermore, the proposed nonlinear PDE controller demonstrates better performance in tracking the desired angular position ($\theta_d$) compared to the conventional PID controller. However, both the conventional PID controller and the proposed nonlinear PDE controller exhibited poor performance in vibration suppression. As a result, this paper proposes a DRL motion planner designed to generate the optimal path with minimal vibrations. The results using the DRL motion planner are presented in the following section.

\subsection{Simulation Results Using the Proposed PDE Controller Combined With the DRL Motion Planner}

As observed in the previous subsection, both control and motion planning are crucial for vibration suppression in flexible manipulators. As explained in the previous chapter, the SAC algorithm is proposed for motion planning, generating the desired angular position ($\theta_d$). 

The DRL agent, as described in Table \ref{DRL Parameters}, operates at a frequency of \(10\) Hz, processing model states every \(0.1\) seconds and generating corresponding actions in the form of desired velocities for the low-level controller. These actions are initially constrained within the range \([-1, 1]\), and then scaled to fit within \([- \dot{\theta}_{\max}, \dot{\theta}_{\max}]\), where \(\dot{\theta}_{\max} = 5\) deg/s. The low-level controller, on the other hand, operates at a significantly higher frequency of \(1\) kHz. Additionally, the reaching reward and the failure penalty, presented in equation (\ref{Reward function}), are defined as follows:
\begin{equation}
\begin{aligned}
    \label{Reward function 2}
    R_{reach} &= +200, \quad \text{if} \quad  
    \begin{cases} 
        |e_T(t)| < 0.1, \\ 
        |\dot{\theta}(t)| < 0.1, \\ 
        |\dot{\omega}(L, t)| < 0.1
    \end{cases},
\end{aligned}
\end{equation}
\begin{equation}
\begin{aligned}
    \label{Reward function 3}
    R_{failure} &= -200, \quad \text{if} \quad  
    \begin{cases} 
        \theta(t) < 0, \\ 
        \text{or} \\ 
        \theta(t) > 90
    \end{cases}.
\end{aligned}
\end{equation}

\textcolor{black}{ 
To promote generalization in the DRL-based motion planner, a random target angle $\theta_T(t)$ and initial angle $\theta(0)$ are selected within the operational range of $[0^\circ, 90^\circ]$ at the start of each episode. Furthermore, to enhance robustness against system variations and bridge the sim-to-real gap, we adopt a domain randomization approach. In this method, key physical parameters of the environment—namely $[L, m, M, I_m, \rho A, EI]$—are perturbed by a random variation of $\pm10\%$ around their nominal values listed in Table~\ref{Parameters}. This exposes the agent to a diverse set of scenarios during training, enabling it to learn policies resilient to uncertainty. Fig.~\ref{Rewards} illustrates the training performance in terms of episode rewards, based on an implementation in MATLAB and Simulink. As shown, the agent's reward converges over time, indicating acceptable learning performance and stable policy development.
\begin{figure}[H]
\centering
\includegraphics[width=0.45\linewidth]{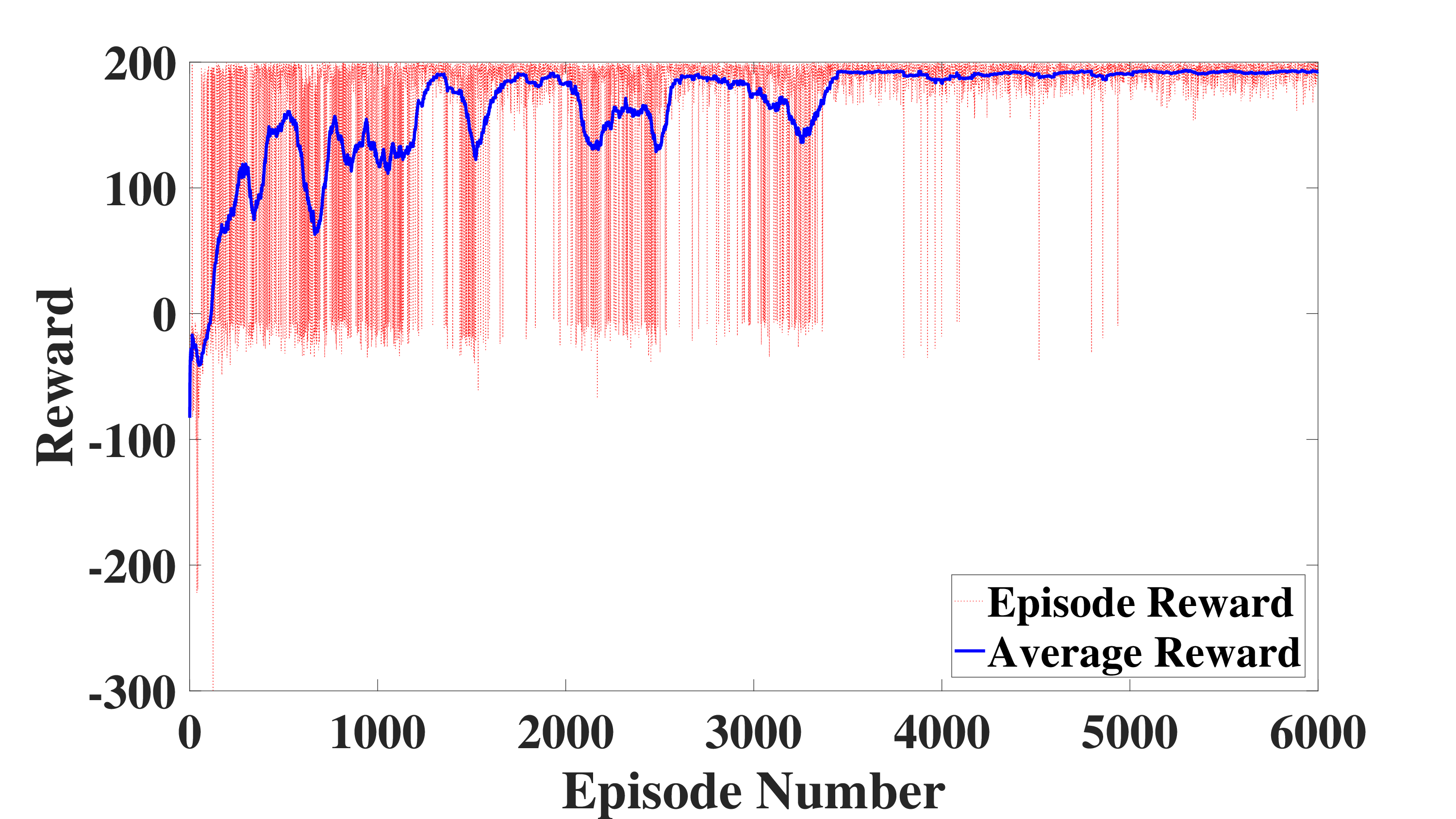}
\caption{\textcolor{black}{Rewards in the Training Process.}}
\label{Rewards}
\end{figure}
}

\begin{figure}[H]
\centering
\subfloat[]{\includegraphics[width=1.3in]{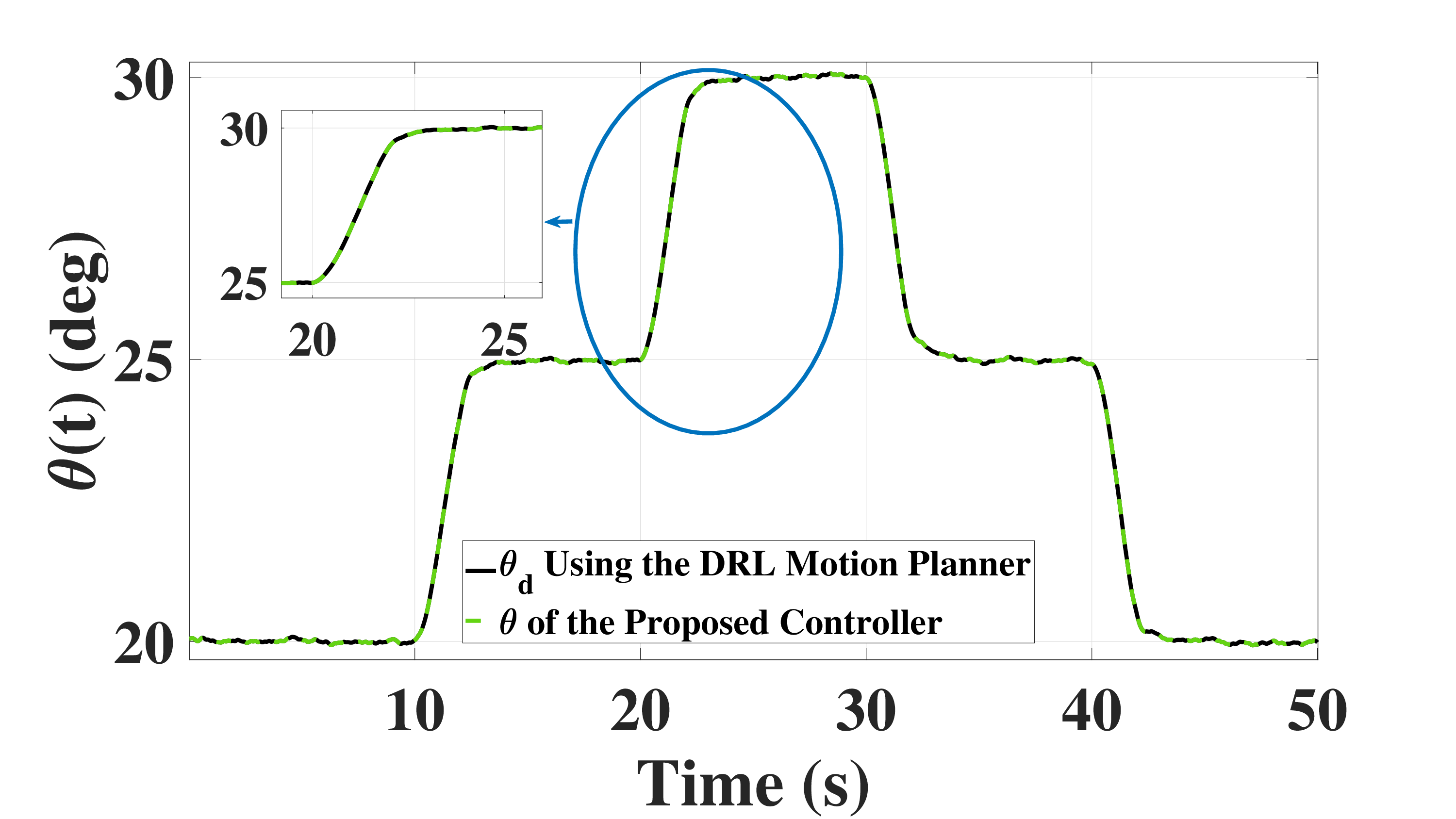}%
\label{Simulation DRL Theta}}

\subfloat[]{\includegraphics[width=1.3in]{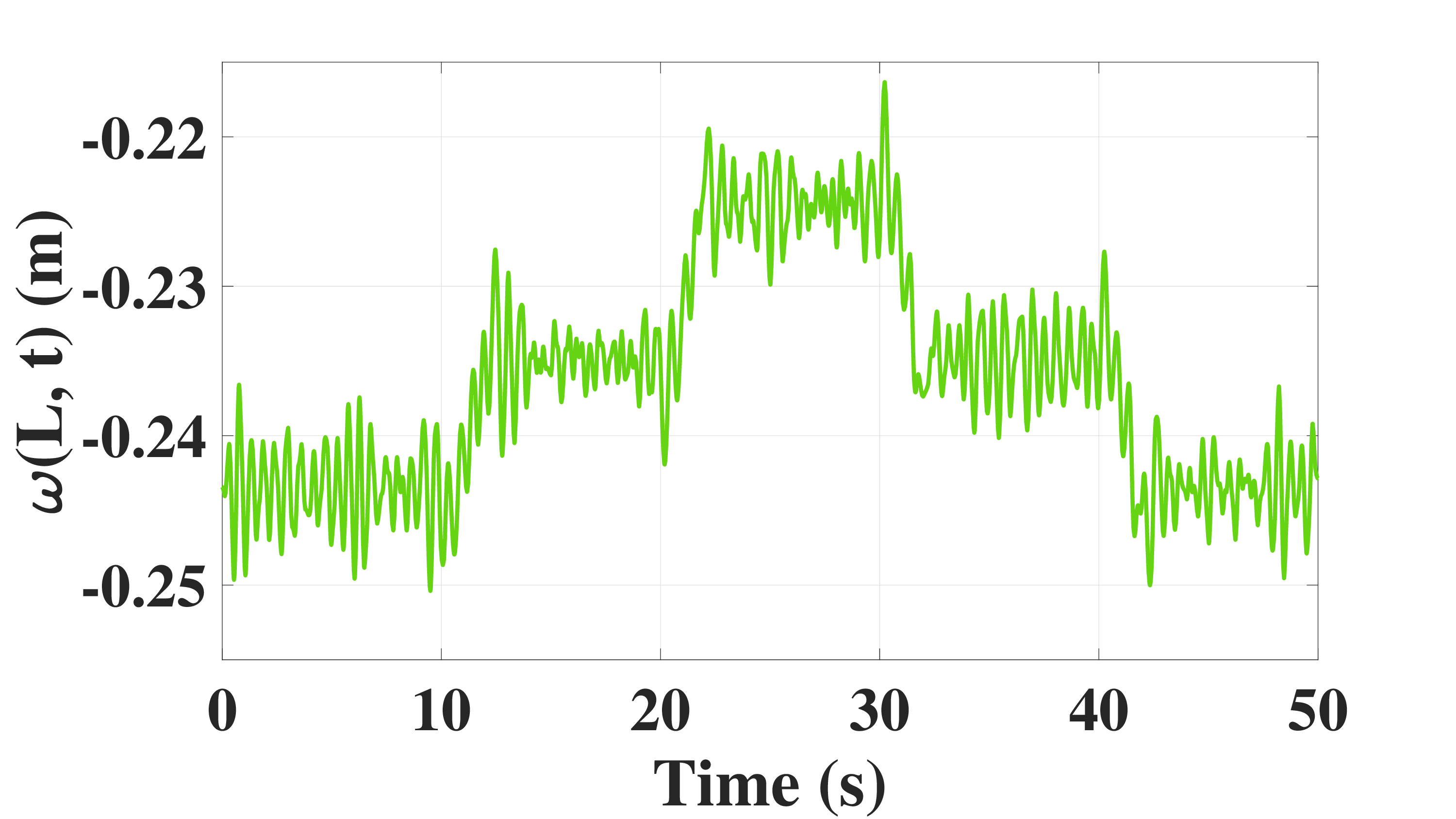}%
\label{Simulation DRL Omega}}
\subfloat[]{\includegraphics[width=1.3in]{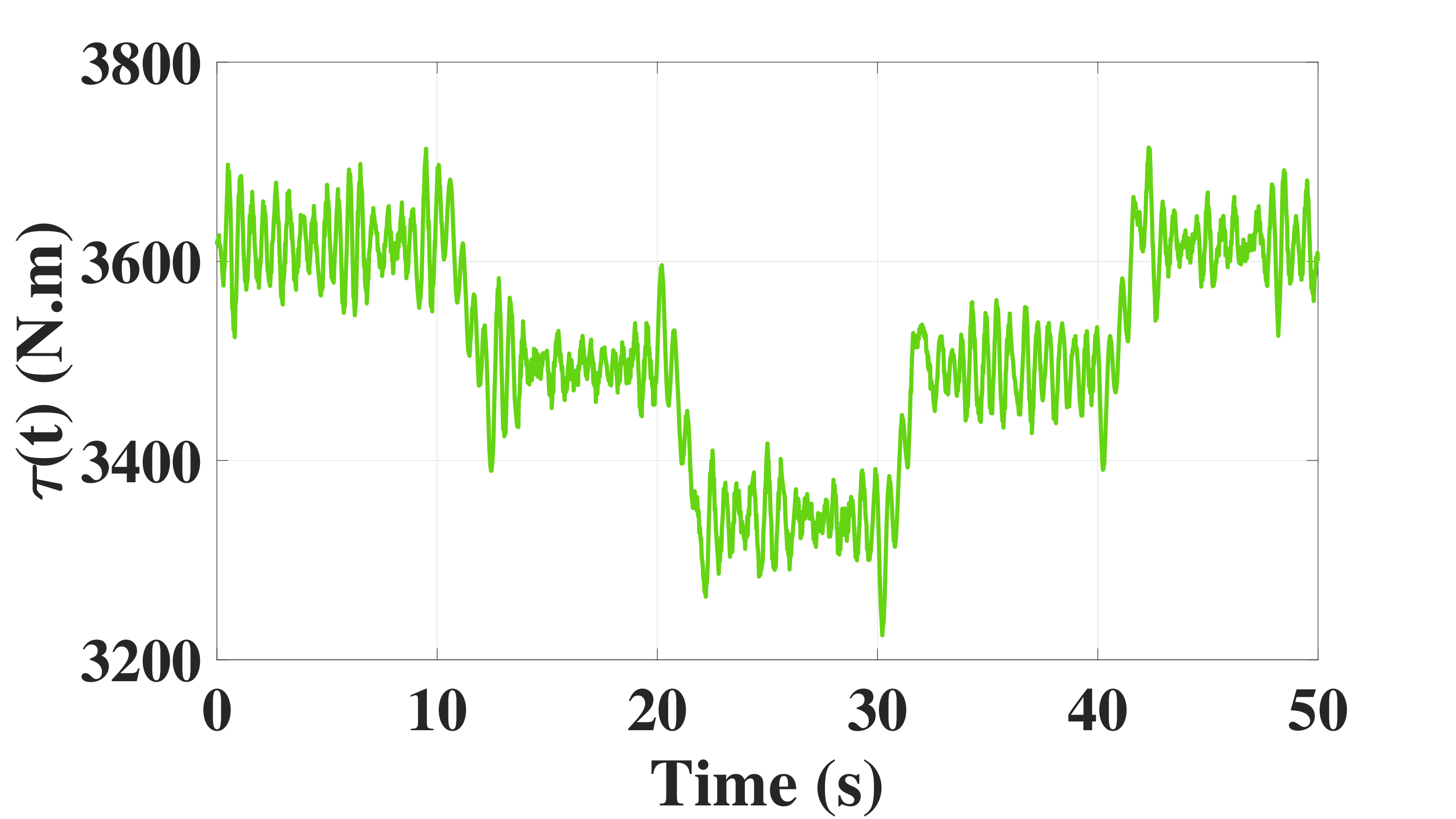}%
\label{Simulation DRL Torque}}
\caption{\textcolor{black}{Simulation Results of the Proposed PDE Controller Using the Proposed DRL Motion Planner. (a) Angular Position. (b) Elastic Deviation of the Payload. (c) Control Input Torque.}}
\label{Simulation DRL}
\end{figure}

The simulation results using the proposed PDE controller and the DRL motion planner are illustrated in Fig. \ref{Simulation DRL}. As seen in Fig. \ref{Simulation DRL Theta}, the proposed controller successfully tracks the desired angular position generated by the proposed motion planner. Furthermore, as shown in Fig. \ref{Simulation DRL Omega}, the proposed DRL motion planner effectively suppresses the vibrations, keeping them under $1.5 \text{ cm}$.

 \textcolor{black}{To evaluate vibration suppression at the flexible link tip, we compared the RMSE of $\dot{\omega}(L, t)$. Our PDE controller with DRL motion planner achieved the lowest RMSE of $1.2906 \times 10^{-4}~\text{m/s}$, compared to $1.6 \times 10^{-3}~\text{m/s}$ for PDE controller with CPT method and $2.0807 \times 10^{-4}~\text{m/s}$ for PID controller with CPT method, confirming its superior vibration mitigation.}
 \begin{table}[ht]
\caption{Parameters of the DRL Agent and their Values}
\centering
\begin{tabular}{|c| c|c|c|}
\hline
\textbf{Parameters} & \textbf{Value} & \textbf{Parameters} & \textbf{Value}\\ [0.5ex]
\hline
Batch size & $128$ & Initial random steps & $500$  \\ \hline
Experience buffer length & $10^6$ & Training episodes & $5000$  \\ \hline
Discount factor & $0.99$ & Max time steps/episode & $300$ \\ \hline
Time step & $0.1$ & $W_e$ & $-5 \times 10^{-3}$  \\ \hline
Learning rate & $10^{-4}$ & $W_{\dot{\theta}}$ & $-10^{-3}$ \\ \hline
Target smoothing factor & $0.001$ & $W_{\dot{\omega}}$ & $-3 \times 10^{-1}$ \\ \hline
\end{tabular}
\label{DRL Parameters}
\end{table}

\section{Experimental Results}
In the previous chapter, it was observed in simulation that the proposed PDE controller combined with the proposed DRL motion planner could effectively track the desired angle while suppressing the endpoint vibrations.

Now, in this chapter, the performance of the proposed controller and motion planner is evaluated on the real flexible robotic manipulator. Therefore, the performance of the proposed PDE controller without and with the proposed DRL motion planner are evaluated in the following sections.

\subsection{Experimental Platform}

\begin{figure}[H]
\centering
\includegraphics[width=0.5\linewidth]{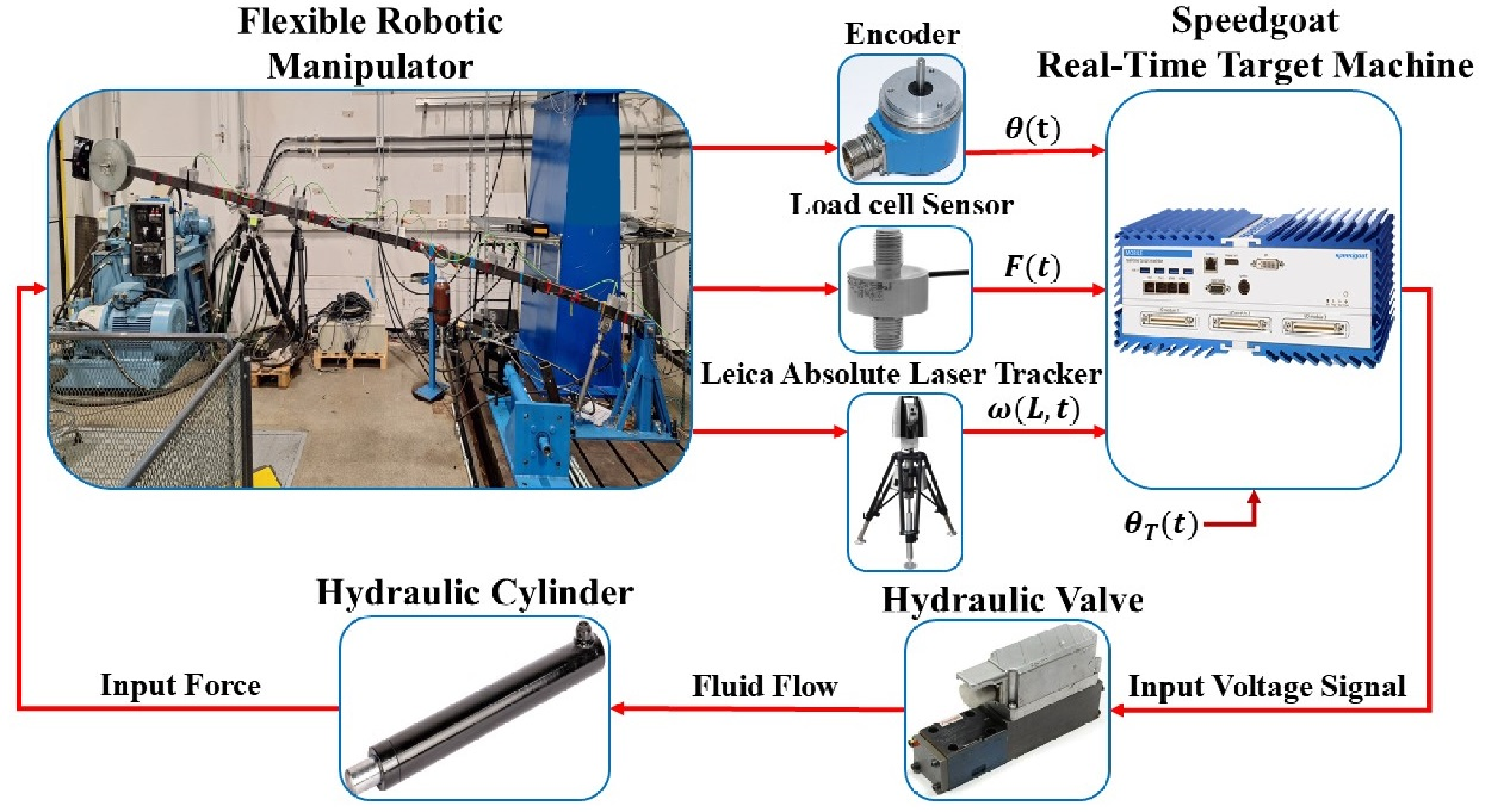}
\caption{Experimental Platform of the Flexible Manipulator.}
\label{Experimental Platform}
\end{figure}

The experimental setup of the flexible robotic manipulator is shown in Fig. \ref{Experimental Platform}.

In general, the robot consists of several components, each serving a specific role, as outlined below:

\subsubsection{Flexible Link}
The flexible link is constructed from OPTIM 700 MH Plus, a high-strength structural steel. With a length of 4.5 m and a cross-sectional area of 60 mm $\times$ 60 mm (thickness of 3 mm), the beam's mass is approximately 22.5kg, while it carries an external load at the end of its length. The steel has a yield strength of 700 MPa, with a tensile strength ranging from 750 to 950 MPa.

\subsubsection{Hydraulic Valve and Cylinder}
For controlling the link's angular position, a hydraulic-actuated system consisting of a $\oslash$35/25-300 mm single-acting hydraulic cylinder and a Bosch Rexroth 4WRPEH servo valve is used.

\subsubsection{Measurement of the Actual Force}
For measuring the amplitude of the force applied by the hydraulic cylinder ($F$), an HBM U9B load cell is utilized. \color{black} As the hydraulic cylinder is attached to the basement, 29 cm from the revolute joint ($r = 29 \text{ cm}$), the torque of $T = r \times F$ is applied to the flexible link that controls the angular position of the robot \color{black}.

\subsubsection{Measurement of the Angular Position}
To measure the angular position of the flexible link ($\theta$), a SICK DS460 Encoder is used.

\subsubsection{Measurement of the Position of the Payload}
For deriving the position of the payload ($y(L,t)$), this project uses the Leica Absolute Laser Tracker AT960-LR, a cutting-edge portable measurement device designed for high-precision tracking of objects in three-dimensional space.

\color{black} 
By knowing the position of the endpoint, using the Laser Tracker, in conjunction with the angular position, using the Encoder, the elastic deviation of the endpoint ($\omega(L, t) = y(L,t) - L\theta(t)$) is calculated. \color{black}

\subsection{Experimental Results Using the Proposed PDE Controller}
In this section, the proposed PDE controller, given by equation (\ref{Tau}), is implemented on the flexible robotic manipulator shown in Fig. \ref{Experimental Platform}. It should be noted that, in this section, the DRL motion planner is not used, and the desired angle ($\theta_d$) is generated using the CPT method. The results are illustrated in Fig. \ref{Experiment Comparison}, represented in red.

As seen in Fig. \ref{Experiment PDE Theta}, the proposed controller effectively tracks the desired angle. However, the endpoint vibrations, as shown in Fig. \ref{Experiment Comparison Omega}, are not acceptable, particularly during lowering.

To evaluate the robustness of the proposed controller, parametric uncertainty should be considered in this research. The main parameters that are used in the proposed controller (\ref{Tau}) are [$E$, $I$, $m$, $L$, $M$]. Therefore, the Root Mean Squared Error (RMSE) of the angular position with uncertainties ranging from $-40\%$ to $+40\%$, compared to the nominal parameters presented in Table \ref{Parameters}, are presented in Table \ref{Uncertainties}.
\begin{table}[H]
\centering
\caption{RSME of the Angular Position, Using the Proposed Controller, with Different Uncertainties}
\begin{tabular}{|c|c|}
\hline
\textbf{Uncertainties} & \textbf{Root Mean Squared Error (Unit)} \\ [0.5ex]
\hline
$-40\%$ Uncertainty & $0.2934 (deg)$  \\ \hline
$-20\%$ Uncertainty & $0.2257 (deg)$  \\ \hline
Nominal Parameters & $0.1875 (deg)$  \\ \hline
$+20\%$ Uncertainty & $0.2041 (deg)$  \\ \hline
$+40\%$ Uncertainty & $0.2712 (deg)$  \\ \hline
\end{tabular}
\label{Uncertainties}
\end{table}

It can be seen that although the controller with nominal parameters has the best performance in tracking the desired angle, with an RMSE of $0.1875 (deg)$, the controllers under uncertainties still exhibit acceptable performance, with a maximum RMSE of $0.2934 (deg)$.

\subsection{Experimental Results Using the Proposed PDE Controller Combined With the Proposed DRL Motion Planner}
\begin{figure}[ht]
\centering
\subfloat[]{\includegraphics[width=1.3in]{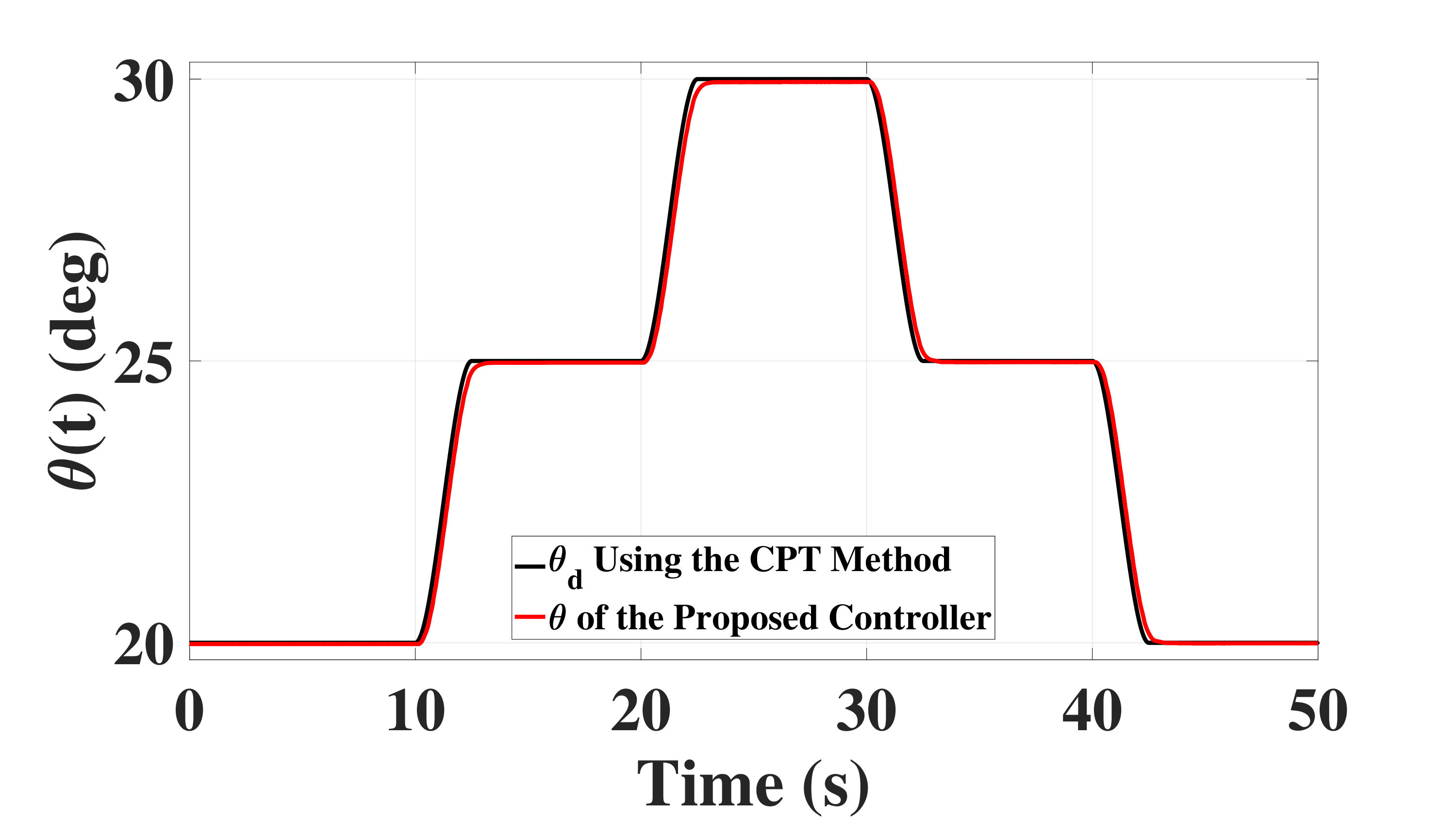}%
\label{Experiment PDE Theta}}
\subfloat[]{\includegraphics[width=1.3in]{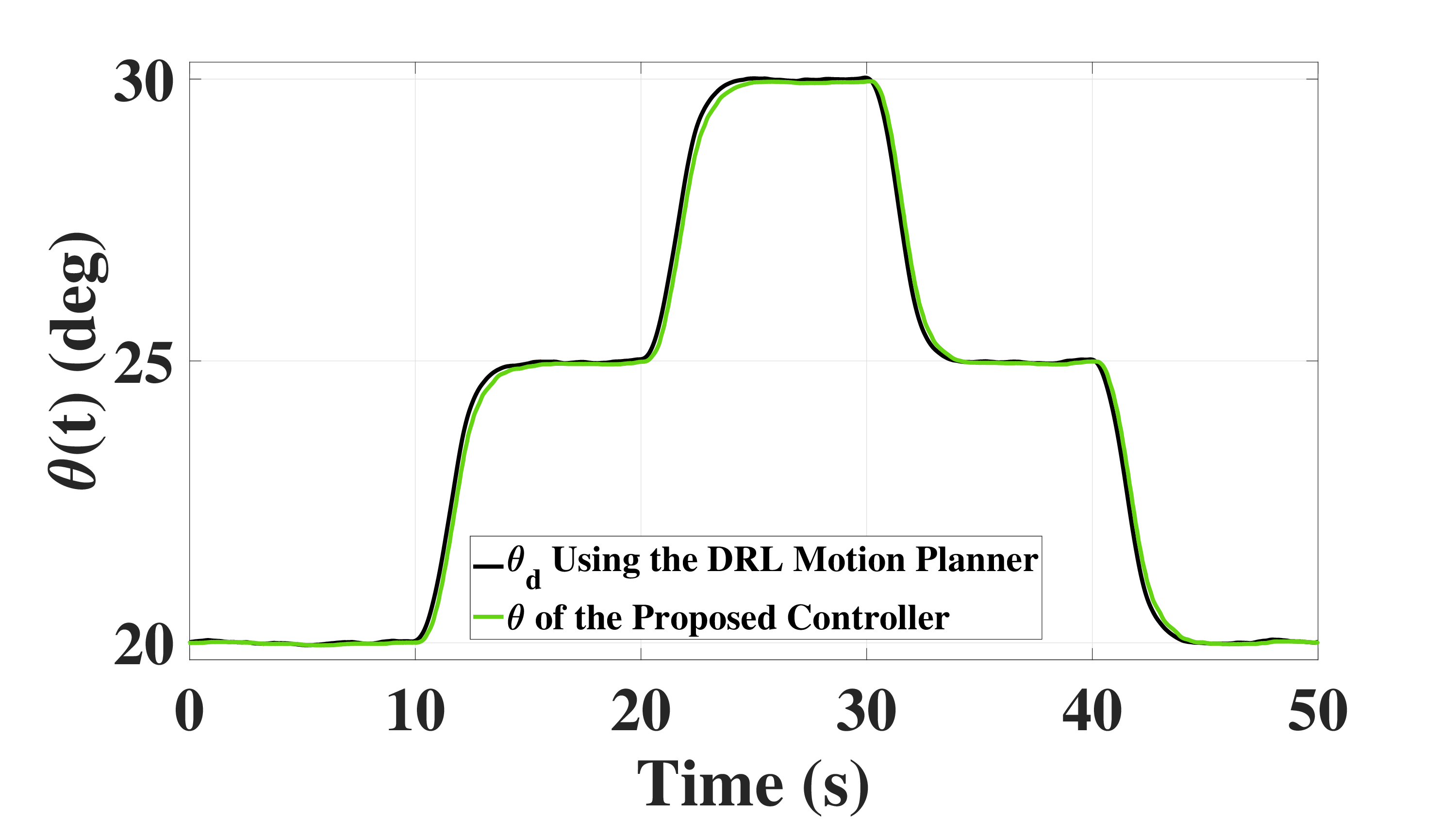}%
\label{Experiment DRL Theta}}

\subfloat[]{\includegraphics[width=1.3in]{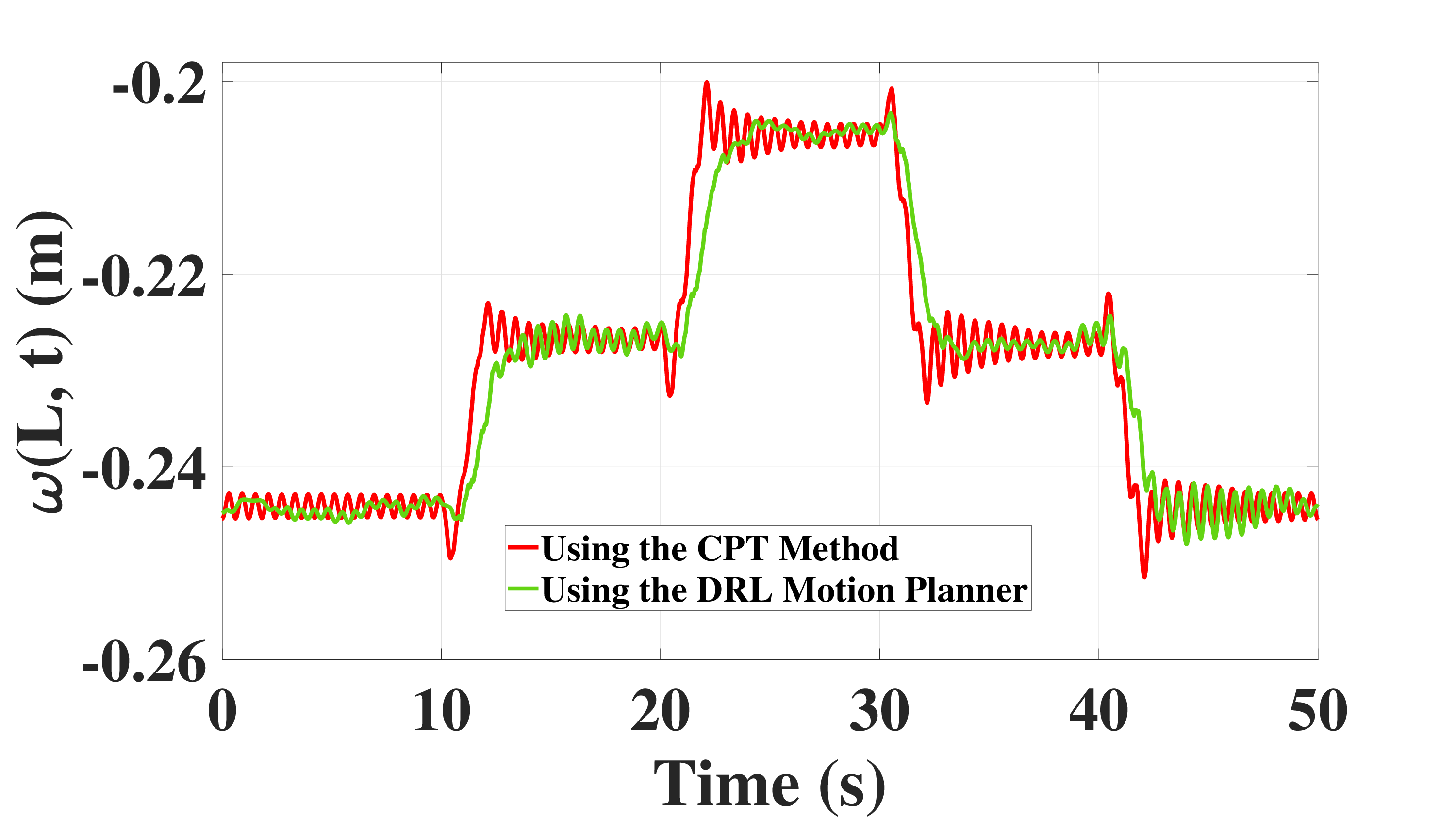}%
\label{Experiment Comparison Omega}}
\subfloat[]{\includegraphics[width=1.3in]{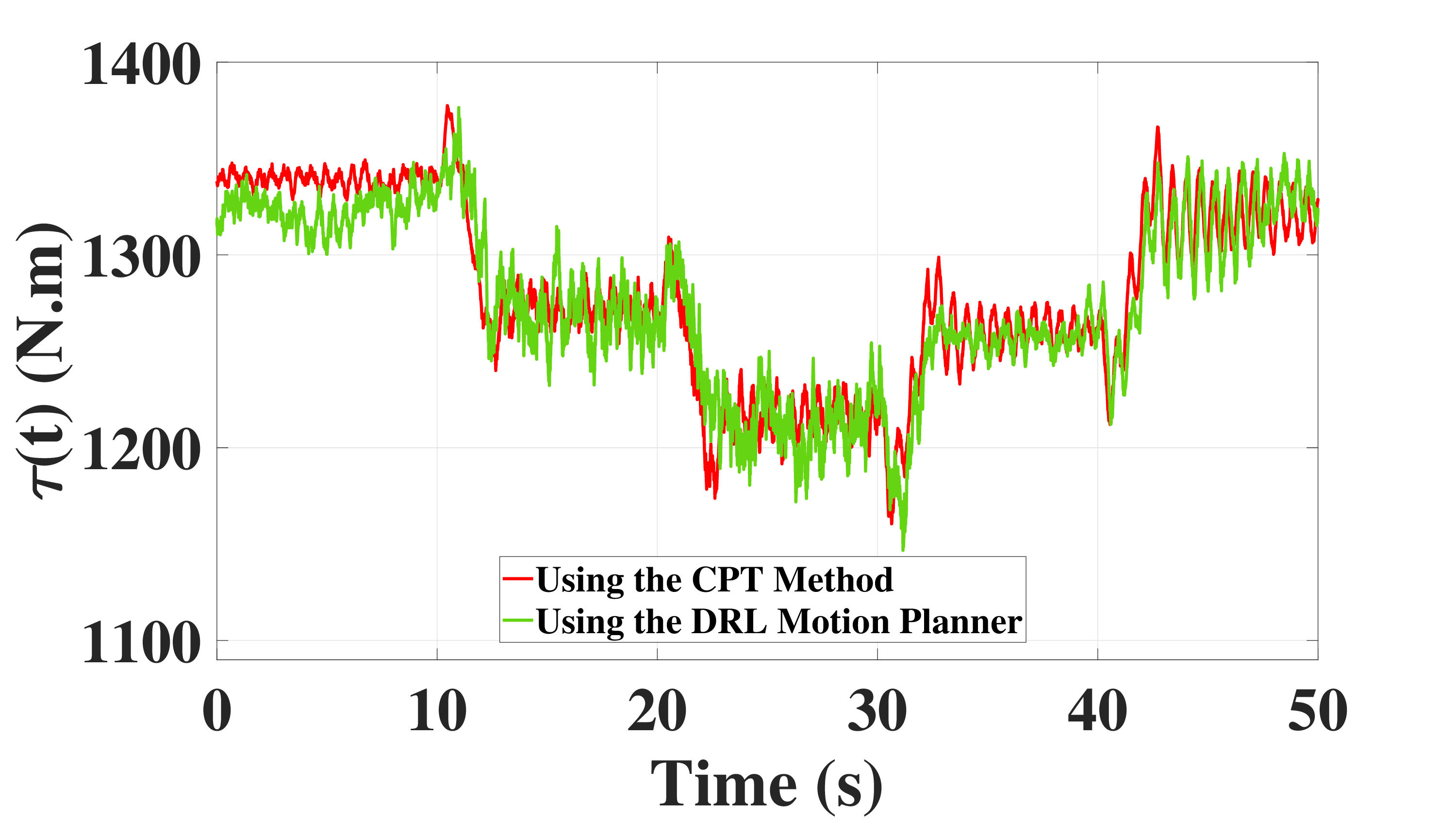}%
\label{Experiment Comparison Torque}}
\caption{\textcolor{black}{Experimental Results. (a) Angular Position Using the CPT Method. (b) Angular Position Using the DRL Motion Planner. (c) Comparison of the Elastic Deviation of the Payload. (d) Comparison of the Control Input Torque.}}
\label{Experiment Comparison}
\end{figure}

In this section, the proposed PDE controller is evaluated when the desired angle ($\theta_d$) is generated using the proposed DRL motion planner. The proposed DRL motion planner is designed to suppress the vibrations and generate the optimal path between each pair of angles. The results are depicted in Fig. \ref{Experiment Comparison}, using green coloring.

\textcolor{black}{As shown in Fig.~\ref{Experiment DRL Theta}, the DRL motion planner produces smooth transitions between angles. Compared to CPT, it achieves superior vibration suppression (Fig.~\ref{Experiment Comparison Omega}), keeping amplitudes below $3~\text{mm}$ when raising and $6~\text{mm}$ when lowering. Notably, unlike model-based methods such as model predictive control which are sensitive to modeling uncertainties, the model-free DRL approach adapts through interaction, enabling reliable trajectory generation despite uncertainties. Moreover, DRL offers superior real-time performance over offline algorithms like CPT, making it well-suited for the fast dynamics of flexible systems.} 

\section{Conclusion}
This paper presented a novel hybrid control approach for flexible robotic manipulators, integrating a nonlinear PDE controller with a DRL motion planner. The primary objective was to achieve precise trajectory tracking while suppressing endpoint vibrations, which is an inherent challenge in flexible manipulators.

Numerical simulations demonstrated that the proposed PDE controller outperforms the PID controller in terms of tracking accuracy and stability; however, it was insufficient in mitigating vibrations due to the system’s underactuated nature. To address this limitation, a DRL motion planner was introduced, leveraging the SAC algorithm \textcolor{black}{with domain randomization}, to generate an optimized trajectory that minimizes vibrations.

Experimental validations confirmed the effectiveness of the proposed approach, that it successfully reduced vibration amplitudes to within $3 \, \text{mm}$ (raising) and $6 \, \text{mm}$ (lowering)—a substantial improvement for a heavy duty flexible manipulator.

While the proposed method significantly enhanced performance, future research may explore:
\begin{itemize}
    \item Enhancing the DRL agent to adapt to varying payloads and dynamic uncertainties.
    \item Extending the DRL framework to optimize not only vibration suppression but also energy efficiency and execution speed.
    \item Implementing the approach on multi-link flexible manipulators to validate its scalability in more complex robotic systems.
    \item \textcolor{black}{Investigating fine-tuning of the DRL motion planner directly on the real system, following initial training in simulation, to further improve real-world adaptability and address unmodeled dynamics.}
\end{itemize}


%






%

\bibliographystyle{IEEEtran}
\bibliography{MyReferences.bib}

\end{document}